\def\paperID{XXXX}
\def\confName{XXXX}
\def\confYear{xxxx}

\def\paperTitle{BEVNeXt: Reviving Dense BEV Frameworks for 3D Object Detection}

\def\authorBlock{
    Zhenxin Li\textsuperscript{1, 2} \quad
    Shiyi Lan\textsuperscript{3} \quad
    Jose M. Alvarez\textsuperscript{3} \quad 
    Zuxuan Wu\textsuperscript{1, 2$\dagger$} \\
  {$^1$Shanghai Key Lab of Intell. Info. Processing, School of CS, Fudan University} \\
{$^2$Shanghai Collaborative Innovation Center of Intelligent Visual Computing} \quad
{$^3$NVIDIA} 
}

\newif\ifreview 
\newif\ifarxiv \newcommand{\arxiv}{\arxivtrue}
\newif\ifcamera 
\newif\ifrebuttal 

\arxiv

\pdfoutput=1
\documentclass[10pt,twocolumn,letterpaper]{article}
\ifreview \usepackage[review]{cvpr} \fi
\ifarxiv \usepackage[pagenumbers]{cvpr} \fi
\ifrebuttal \usepackage[rebuttal]{cvpr} \fi
\ifcamera \usepackage{cvpr} \fi

\usepackage{graphicx}	
\usepackage{amsmath}	
\usepackage{amssymb}	
\usepackage{booktabs}
\usepackage{times}
\usepackage{microtype}
\usepackage{epsfig}
\usepackage[table,xcdraw,dvipsnames]{xcolor}
\usepackage{caption}
\usepackage{float}
\usepackage{placeins}
\usepackage{color, colortbl}
\usepackage{stfloats}
\usepackage{enumitem}
\usepackage{tabularx}
\usepackage{xstring}
\usepackage{multirow}
\usepackage{xspace}
\usepackage{url}
\usepackage{subcaption}
\usepackage{xcolor}
\usepackage[hang,flushmargin]{footmisc}
\usepackage{bm}

\usepackage[accsupp]{axessibility}

% ...

\ifarxiv  \fi

\newcommand{\R}[1]{{%
    \textbf{%
        \ifstrequal{#1}{1}{\textcolor{red}{R#1}}{%
        \ifstrequal{#1}{2}{\textcolor{blue}{R#1}}{%
        \ifstrequal{#1}{3}{\textcolor{magenta}{R#1}}{%
        \ifstrequal{#1}{4}{\textcolor{teal}{R#1}}{%
                           \textcolor{cyan}{R#1}%
        }}}}%
    }%
}}

\newcommand\blfootnote[1]{%
  \begingroup
  \renewcommand\thefootnote{}\footnote{#1}%
  \addtocounter{footnote}{-1}%
  \endgroup
}  

\usepackage{xr-hyper}

\makeatletter
\newcommand*{\addFileDependency}[1]{
  \typeout{(#1)}
  \@addtofilelist{#1}
  \IfFileExists{#1}{}{\typeout{No file #1.}}
}

\makeatother

\definecolor{cvprblue}{rgb}{0.21,0.49,0.74}
\usepackage[pagebackref,breaklinks,colorlinks,citecolor=cvprblue]{hyperref}
\usepackage[capitalize]{cleveref}
\crefname{section}{Sec.}{Secs.}
\crefname{table}{Table}{Tables}
\crefname{figure}{Fig.}{Figs.}

\usepackage{indentfirst} 
\usepackage{pgfplots}

\makeatletter
\let\@afterindentfalse\@afterindenttrue
\makeatother

\begin{document}

\title{\paperTitle}
\author{\authorBlock}

\maketitle

\begin{abstract}
Recently, the rise of query-based Transformer decoders is reshaping camera-based 3D object detection. These query-based decoders are surpassing the traditional dense BEV (Bird's Eye View)-based methods. 
However, we argue that dense BEV frameworks remain important due to their outstanding abilities in depth estimation and object localization, depicting 3D scenes accurately and comprehensively. 
This paper aims to address the drawbacks of the existing dense BEV-based 3D object detectors by introducing our proposed enhanced components, including a CRF-modulated depth estimation module enforcing object-level consistencies, a long-term temporal aggregation module with extended receptive fields, and a two-stage object decoder combining perspective techniques with CRF-modulated depth embedding. These enhancements lead to a ``modernized'' dense BEV framework dubbed BEVNeXt. 
On the nuScenes benchmark, BEVNeXt outperforms both BEV-based and query-based frameworks under various settings, achieving a state-of-the-art result of 64.2 NDS on the nuScenes test set. Code will be available at \url{https://github.com/woxihuanjiangguo/BEVNeXt}.
\blfootnote{$^{\dagger}$Corresponding author.}
\end{abstract}

\section{Introduction}
\label{sec:intro}

Visual-based 3D object detection~\cite{huang2021bevdet, liu2022petr, wang2022detr3d, li2022bevformer} is a critical component of autonomous driving and intelligent transportation systems. 
Unlike LiDAR-based systems with access to depth data, the primary challenge in visual-based 3D object detection is accurately perceiving depth, a task largely reliant on empirical knowledge of images.
As an important part of detection, object localization depends heavily on the accuracy of depth~\cite{park2022time}.
Precise and robust object localization is the cornerstone of 3D perception, as it helps identify obstacles~\cite{badrloo2022image}, lays a foundation for scene forecasting~\cite{ngiam2021scene, zhou2023query}, and leads to reassuring planning~\cite{hu2023planning, teng2023motion}.

\begin{figure}
    \centering
        \centering
        \includegraphics[width=\linewidth]{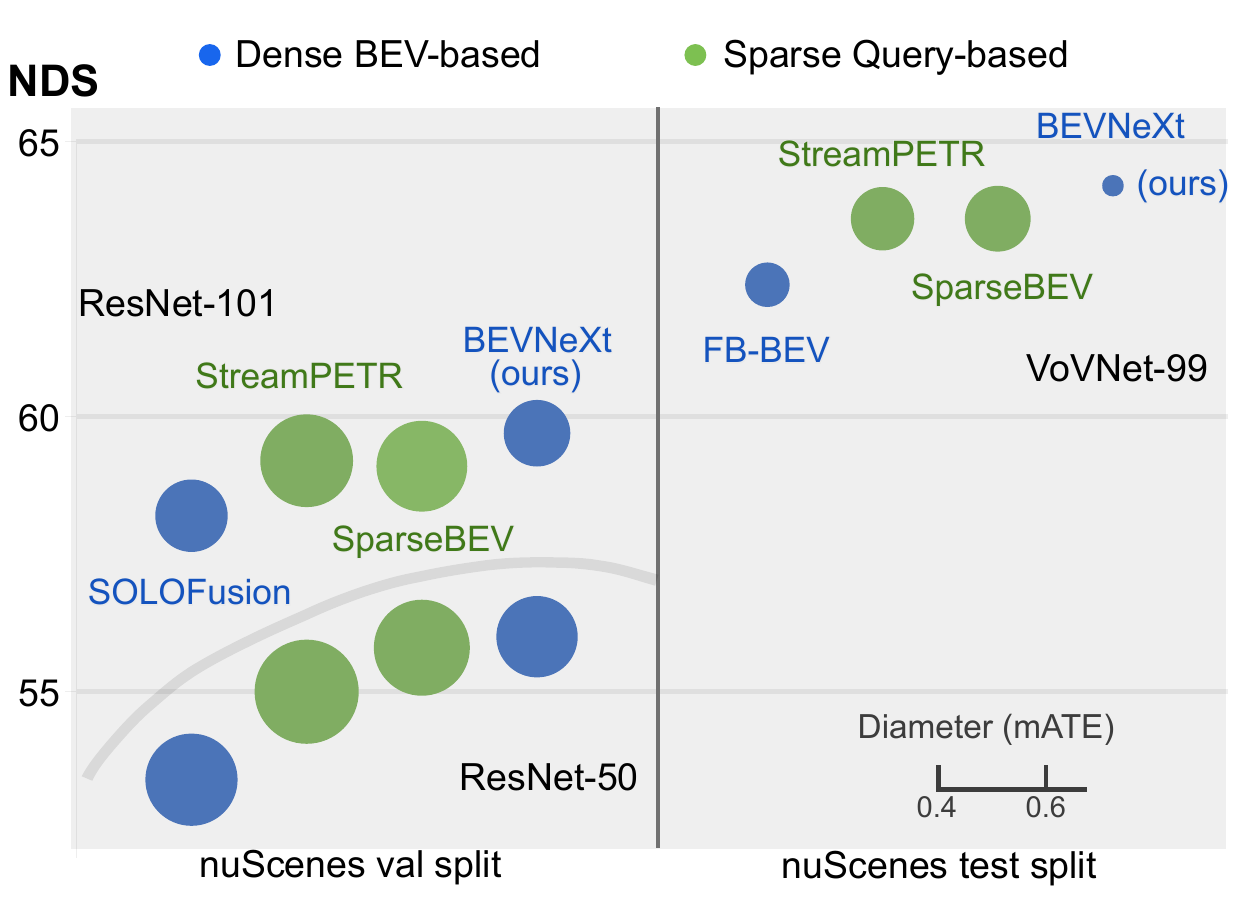}
    \caption{\textbf{Previous SOTAs vs.\ BEVNeXt on the nuScenes 3D Object Detection Benchmark.} On the nuScenes val split and test split, we compare BEVNeXt with previous SOTAs using ($\text{ResNet-50, }$bottom in the left panel), ($\text{ResNet-101, }$top in the left panel), and ($\text{VoVNet-99, }$right panel) as the backbone. 
    BEVNeXt outperforms all previous sparse query-based ones in terms of comprehensive performance, meanwhile generating much fewer localization errors. The diameter of each bubble represents the mean Average Translation Error (mATE) each model produces. Higher and smaller bubbles are better. Best viewed in color.}
\label{fig:header}
\end{figure}
To detect 3D objects with visual information, two research directions have prevailed:
\textit{dense BEV (Bird's Eye View)-based methods} and \textit{sparse query-based methods}. BEV-based methods transform image feature maps into one unified dense bird's-eye-view feature map and thus apply detection decoders on it. In contrast, sparse query-based methods learn a set of object queries, which focus on sparse foreground objects rather than background details, and then predict 3D objects by leveraging multiple stages of cross-attention between object queries and image features and self-attention among object queries.

Despite the superior performance of recent query-based methods over dense BEV-based approaches, we maintain that retaining the dense feature map is advantageous for a complete environmental understanding, regardless of background or foreground elements. 
This trait makes BEV-based frameworks suitable for dense prediction tasks such as occupancy prediction~\cite{tong2023scene, tian2023occ3d}.
Further, the dense processing equips them with robustness in object localization, which accounts for their fewer localization errors compared with sparse counterparts as shown in Fig.~\ref{fig:header}.
We argue that BEV-based detectors lag behind query-based ones due to less advanced network designs and training techniques. Building on this, we summarize the shortcomings of classic dense BEV-based approaches as follows:

\begin{itemize}
    
    \vspace{0.05in}\item \textbf{Insufficient 2D Modeling.} 
    Recent sparse query-based methods have demonstrated that improved 2D modeling can significantly enhance detection accuracy~\cite{wang2022focal, jiang2023far3d}. In dense BEV-based approaches, efforts to boost 2D modeling include an auxiliary depth estimation task supervised by LiDAR inputs~\cite{li2023bevdepth}. Yet, the impact is constrained by the low resolution of LiDAR points~\cite{caesar2020nuscenes}, leading to imprecise depth perception and suboptimal model performance.
   
    \vspace{0.05in}\item \textbf{Inadequate Temporal Modeling.} BEV frameworks often suffer from limited temporal modeling capabilities, which are critical in visual-based 3D detectors~\cite{park2022time, wang2023streampetr, lin2023sparse4d, liu2023sparsebev}. Establishing a large receptive field in a dynamic 3D space during temporal fusion is crucial, especially when the ego vehicle and surrounding objects are in motion. Query-based methods~\cite{wang2023streampetr, liu2023sparsebev} can easily achieve this through the global attention mechanism~\cite{vaswani2017attention}, while BEV-based ones~\cite{park2022time, han2023exploring} are bounded by the locality of convolutions. 

    \vspace{0.05in}\item \textbf{Feature Distortion in Uplifting.} In dense BEV-based methods, feature distortion is a natural consequence of transforming feature maps across different coordinate systems and resolutions. 
    On the other hand, sparse query-based approaches are unaffected since they attend to image feature maps in the 2D space rather than transformed features, thus avoiding feature distortion.

\end{itemize}

We introduce BEVNeXt, a modern dense BEV framework for 3D object detection comprising three main components. 
First, we employ a Conditional Random Field (CRF) to enhance depth accuracy and address depth supervision challenges, integrating depth probabilities with color information without extra supervision or significant computational cost. 
Second, the Res2Fusion module, inspired by Res2Net convolution blocks, expands the receptive field in dynamic 3D settings. 
Third, leveraging the predicted depth information, we have developed a two-stage object decoder. This decoder blends the spirit of sparse query-based techniques with CRF-enhanced depth embedding to improve instance-level BEV features using depth-focused 2D semantics. Together, these elements make BEVNeXt a stronger framework for object detection and localization.

We conduct in-depth experiments on the nuScenes dataset. As shown in Fig.~\ref{fig:header}, BEVNeXt achieves the highest 56.0\% NDS and 64.2\% NDS on the val split and test split, respectively, as well as the lowest mATE compared with all prior methods, demonstrating its outstanding comprehensive performance and preciseness in 3D object localization. More specifically, BEVNeXt outperforms previous state-of-the-art BEV-based SOLOFusion~\cite{park2022time} by 2.6\% and 2.3\% on val split and test split, respectively.

\section{Related Work}
\label{sec:related}

\subsection{Dense BEV-based 3D Object Detection}
\label{subsec:dense bev}

Ever since the pioneering work of LSS~\cite{philion2020lift}, which introduces the concept of lifting 2D image features to the BEV space by predicting pixel-level depth probabilities, a significant research direction~\cite{huang2021bevdet, huang2022bevdet4d, li2023bevdepth, li2022bevstereo, park2022time, han2023exploring, huang2022bevpoolv2} has emerged, dedicated to improving the quality and efficiency of constructing the BEV space for 3D object detection and other perception tasks (\eg map segmentation~\cite{philion2020lift, ng2020bevseg, hu2021fiery}, occupancy prediction~\cite{tian2023occ3d, tong2023scene}). The lifting procedure is also known as forward projection~\cite{li2023fbbev}. In particular, the BEVDet series~\cite{huang2021bevdet, huang2022bevdet4d, huang2022bevpoolv2} proposed an efficient pipeline to perform 3D object detection in the BEV space, as well as the short-term temporal modeling for velocity estimation~\cite{huang2022bevdet4d}. BEVDepth~\cite{li2023bevdepth} and BEVStereo~\cite{li2022bevstereo} have respectively advanced the critical depth estimation process by leveraging explicit supervision from LiDAR point clouds~\cite{li2023bevdepth} and stereo matching~\cite{li2022bevstereo}. To extend their capabilities with long-term temporal information, SOLOFusion~\cite{park2022time} adopts a straightforward concatenation technique across historical BEV representations, demonstrating exceptional performance, while VideoBEV~\cite{han2023exploring} alleviates the heavy computation budget of SOLOFusion with recurrent modelling. However, their long-term fusion strategies suffer from an insufficient receptive field and rely on ego-motion transformation to discriminate stationary objects from moving ones, which can lead to motion misalignment of dynamic objects~\cite{huang2023leveraging}. We argue that expanding the receptive field allows the model to distinguish different objects automatically.

Backward projection~\cite{li2022bevformer, yang2023bevformer} is the inverse operation of forward projection, a technique that samples multi-view 2D features and populates the BEV space with them. In recent advancements presented in FB-BEV~\cite{li2023fbbev} and FB-OCC~\cite{li2023fbocc}, these two projection techniques are unified to obtain a stronger BEV representation, benefiting 3D object detection and occupancy prediction.
This technique is used in our object decoder. 
However, unlike prior work, we utilize backward projection only to refine object-level BEV features, rather than the entire BEV representation. 
Further, this process is boosted with CRF-modulated depth embedding, which proves conducive to attribute prediction.

\subsection{Sparse Query-based 3D Object Detection}
\label{subsec:sparse query}

Following query-based 2D object detectors~\cite{carion2020end, zhu2020deformable}, a parallel avenue of research~\cite{wang2022detr3d, lin2022sparse4d, liu2022petr} has emerged. This alternative approach performs 3D object detection by directly querying 2D features, sidestepping the need for explicit 3D space construction. The querying procedure is typically carried out using the conventional attention mechanism~\cite{vaswani2017attention}, as seen in the PETR series~\cite{liu2022petr, liu2022petrv2, wang2022focal, wang2023streampetr} or the sparse deformable attention mechanism~\cite{zhu2020deformable}, as seen in the Sparse4D series~\cite{lin2022sparse4d, lin2023sparse4d}. After the emergence of SOLOFusion~\cite{park2022time}, the PETR series~\cite{liu2022petr, liu2022petrv2, wang2022focal} embraces the concept of long-term temporal fusion and integrate it into the query-based framework. Through a well-designed propagation algorithm in the query space, StreamPETR~\cite{wang2023streampetr} achieves substantial improvements compared to its static~\cite{liu2022petr, wang2022focal} or short-term counterparts~\cite{liu2022petrv2}. In addition, the recent breakthroughs made by SparseBEV~\cite{liu2023sparsebev} have demonstrated that these object queries can be explicitly defined within the BEV space, while Far3D~\cite{jiang2023far3d} constructs 3D queries through the employment of a 2D object detector and a depth network, significantly expanding the range of 3D object detection. 
These detectors tend to produce more localization errors as they locate objects through the cross-attention mechanism instead of depth information.
Unlike these methods, our work fully builds upon dense BEV frameworks, which are generally more robust in object localization.
\begin{figure*}[tp]
    \centering
    \includegraphics[width=\linewidth]{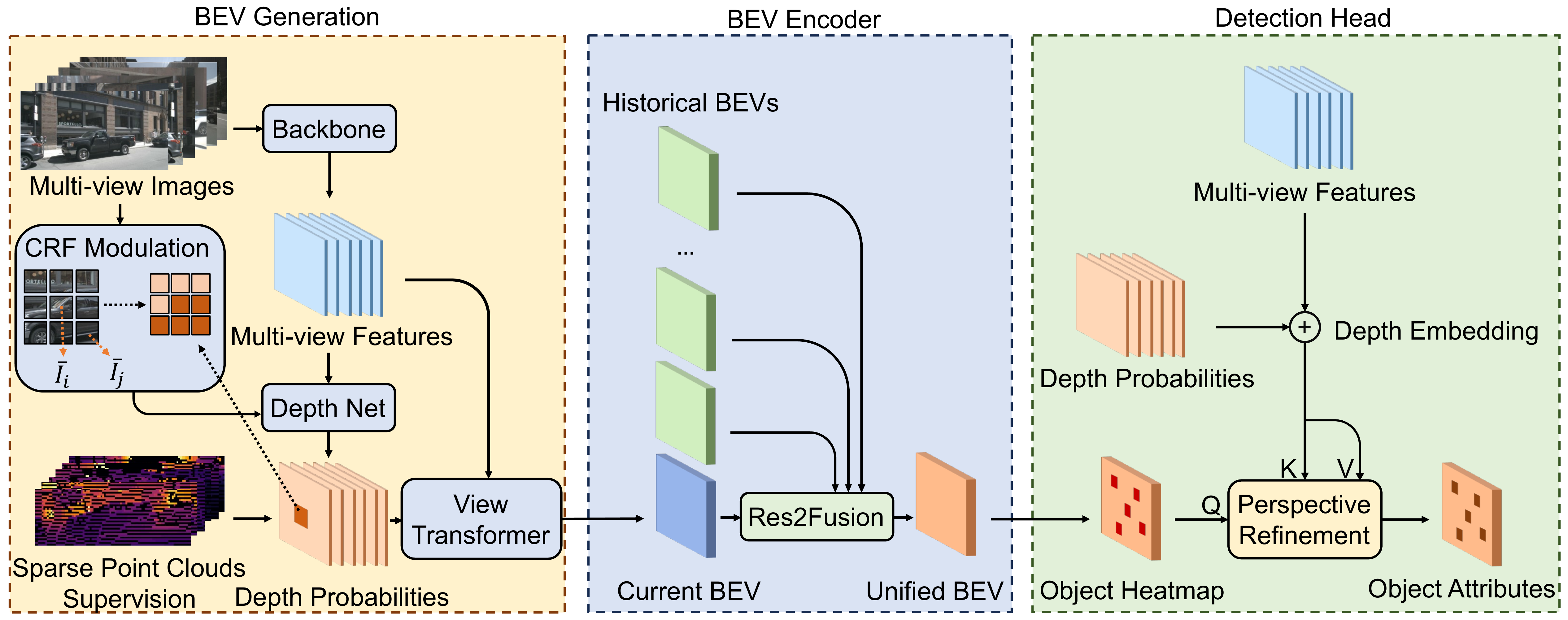}
    \caption{\textbf{Overall Architecture of BEVNeXt.} The backbone first extracts multi-view image features, which are converted into depth distributions with a depth network and CRF modulation. The BEV feature at the current frame is fused with previous ones through a Res2Fusion module. Finally, a CenterPoint detection head, coupled with perspective refinement, generates object heatmaps and attributes.}
    \vspace{-0.2in}

    \label{fig:arch}
\end{figure*}

\subsection{CRF for Dense Predictions}

Conditional Random Fields (CRF) have long been a fundamental tool for addressing dense prediction tasks such as semantic segmentation~\cite{krahenbuhl2011efficient} and depth estimation~\cite{liu2014discrete}, predating the widespread adoption of CNNs. 
With the emergence of CNNs, CRF as RNN~\cite{zheng2015conditional} first shows that CRF can evolve as a seamless part of a CNN, whose job is to modulate pixel-level probabilities generated by the last CNN layer based on observed image features.
For depth estimation, \cite{cao2017estimating} treats the task as a pixel-wise classification problem, making it suitable for the application of CRF, while~\cite{liu2015deep} calculates the CRF energy from continuous depth values. 
Additionally, in the domain of weakly-supervised instance segmentation (WSIS), where pixel-level mask annotations are not available, \cite{lan2021discobox, lan2023vision} leverages CRF to enforce prediction consistency among pixels with similar color characteristics. 

\subsection{3D Object Detection with LiDAR sensors}
LiDAR sensors are widely used in 3D object detection~\cite{yin2021center, shi2022pillarnet, zhou2022centerformer, sun2022swformer, fan2022embracing, chen2023voxelnext} since they generate reliable range information. These detectors typically decode objects from densely encoded BEV features~\cite{yin2021center, fan2022embracing, shi2022pillarnet, zhou2022centerformer} or from sparse voxels~\cite{sun2022swformer, chen2023voxelnext}. Furthermore, LiDAR sensors are jointly used with camera sensors in multi-modal 3D object detectors~\cite{liu2023bevfusion, bai2022transfusion, jiao2023msmdfusion, yang2022deepinteraction, yan2023cross} due to their complementary nature. 
The advancements in this field are often directly applied to the pseudo point clouds generated by dense BEV frameworks~\cite{huang2021bevdet, huang2022bevdet4d} for object decoding. Similarly, our object decoder derives inspiration from CenterFormer~\cite{zhou2022centerformer} and TransFusion~\cite{bai2022transfusion}, two 2-stage center-based detectors. Instead of densely attending to image features~\cite{bai2022transfusion} or BEV features~\cite{zhou2022centerformer}, our decoder employs depth-guided perspective refinement, an enhanced spatial cross-attention mechanism based on BEVFormer~\cite{li2022bevformer}.

\section{Method}
\label{sec:method}
We propose BEVNeXt, an enhanced dense BEV framework building upon existing LSS-based methods.
BEVNeXt consists of three key components as shown in Fig.~\ref{fig:arch}:
\begin{itemize}
  \setlength\itemsep{0.05in}
    \item \textbf{BEV Generation:} Given multi-view images $\{\bm{I^i}\}_{i=1}^6$, the backbone extracts multi-scale image features denoted as $\{F_{1/4}^i, F_{1/8}^i, F_{1/16}^i, F_{1/32}^i\}_{i=1}^6$, which are processed by a depth network for depth probabilities ${\{\boldsymbol{d^i}\}}_{i=1}^6$. For the spatial accuracy of the BEV feature, the CRF layer is employed to modulate ${\{\boldsymbol{d^i}\}}_{i=1}^6$ with image color information $\{\bm{I^i}\}_{i=1}^6$, producing depth probabilities $\{\boldsymbol{\tilde{d^i}}\}_{i=1}^6$ that are consistent on the object level. Then, the View Transformer computes the BEV feature $B_t$ at the current timestamp $t$ using features and modulated depth probabilities.

    \item \textbf{BEV Encoder:} The BEV Encoder is designed to fuse historical BEV features $\{B_{t-k+1}, ..., B_t\}$ across $k$ frames into a unified BEV representation $\tilde{B}$. The aggregation process demands a sufficient receptive field over the dynamic 3D environment, which is fulfilled in Res2Fusion.

    \item \textbf{Detection Head:} Finally, a center-based 3D object detection head~\cite{yin2021center} processes the output $\tilde{B}$ of the BEV Encoder, decoding the BEV representation into 3D objects. CRF-modulated depth probabilities $\{\boldsymbol{\tilde{d^i}}\}_{i=1}^6$ are used as an embedding to help the object decoder attend to discriminative 2D features.
\end{itemize} \leavevmode
For the rest of this section, we elaborate on the specific enhancements of these components following the order of the detection pipeline.

\subsection{CRF-modulated Depth Estimation}
\label{subsec:crf}
In dense BEV-based methods, depth estimation acts as a 2D auxiliary task, improving 2D modeling and potentially helping prevent feature distortion during uplifting. Therefore, obtaining accurate and higher-resolution depth prediction is beneficial. Considering depth estimation as a segmentation task, with each class representing a specific depth range, we can use Conditional Random Fields (CRF) to enhance the depth estimation quality.
Following the utilization of CRFs in sparsely supervised prediction tasks~\cite{liu2015deep, lan2021discobox, lan2023vision}, we aim at using CRF-modulated depth estimation to mitigate insufficient depth supervision by imposing a color smoothness prior~\cite{krahenbuhl2011efficient}, which enforces depth consistency at the pixel level. 
Let $\{X_1, ..., X_N\}$ represent the $N$ pixels in the downsampled feature map $F^i_{1/n}$, acquired by a stride of $n$, and $\{D_1, ..., D_k\}$ be $k$ discrete depth bins. The depth network's responsibility is to assign each pixel to various depth bins, represented as $\boldsymbol{d}=\{x_1,...,x_N|x_i\in\{D_1, ..., D_k\}\}$. The camera index is discarded for convenience. Given this assignment $\boldsymbol{d}$, our objective is to minimize its corresponding energy cost $E(\boldsymbol{d}|\boldsymbol{I})$, as defined following~\cite{krahenbuhl2011efficient}:
\begin{equation}
    E(\boldsymbol{d}|\boldsymbol{I})=\sum_{i}\psi_u(x_i) + \sum_{i\neq j}\psi_p(x_i,x_j),
\end{equation} 
where $\sum_{i}\psi_u(x_i)$ are the unary potentials, measuring the cost associated with the initial output from the depth network.
Building upon prior research~\cite{cao2017estimating, zheng2015conditional}, we define the pairwise potential as:\begin{equation}
    \psi_p(x_i,x_j)=\sum_{w}w\exp{(-\frac{|\boldsymbol{\bar{I}}_i-\boldsymbol{\bar{I}}_j|^2}{2\theta^2})}|x_i-x_j|,
\end{equation} where $\boldsymbol{\bar{I}}_i$ and $\boldsymbol{\bar{I}}_j$ represent average RGB color values of image patches with dimensions matching the downsampling stride. Furthermore, $|x_i-x_j|$ is the label compatibility between two depth bins, which measures the distance of their centers in the real-world scale. The CRF is attached to the depth network as an extra layer. 
We denote the CRF-modulated depth probabilities as $\boldsymbol{\tilde{d}}$.

In existing BEV-based solutions that rely on explicit depth supervision, such as BEVDepth~\cite{li2023bevdepth} and SOLOFusion~\cite{park2022time}, the depth network typically processes the feature map at a small scale (\ie $F_{1/16}^i$). Given a low-resolution input, there is a dense depth label coverage, which comes at the price of discarding too many labels with aggressive downsampling. This, in turn, compromises the effectiveness of our CRF-modulated depth estimation, as indicated in Tab.~\ref{tab:abl_crf}. To demonstrate the effectiveness of our approach under this resolution constraint, our depth network operates on a larger feature map (\ie $F_{1/8}^i$) while halving the channel size. Nevertheless, the advantages of CRF modulation becomes increasingly noticeable as the input resolution scales up, as shown in Tab.~\ref{tab:abl_crf}.

\subsection{Res2Fusion}
\begin{figure}[tp]
    \centering
    \includegraphics[width=\linewidth]{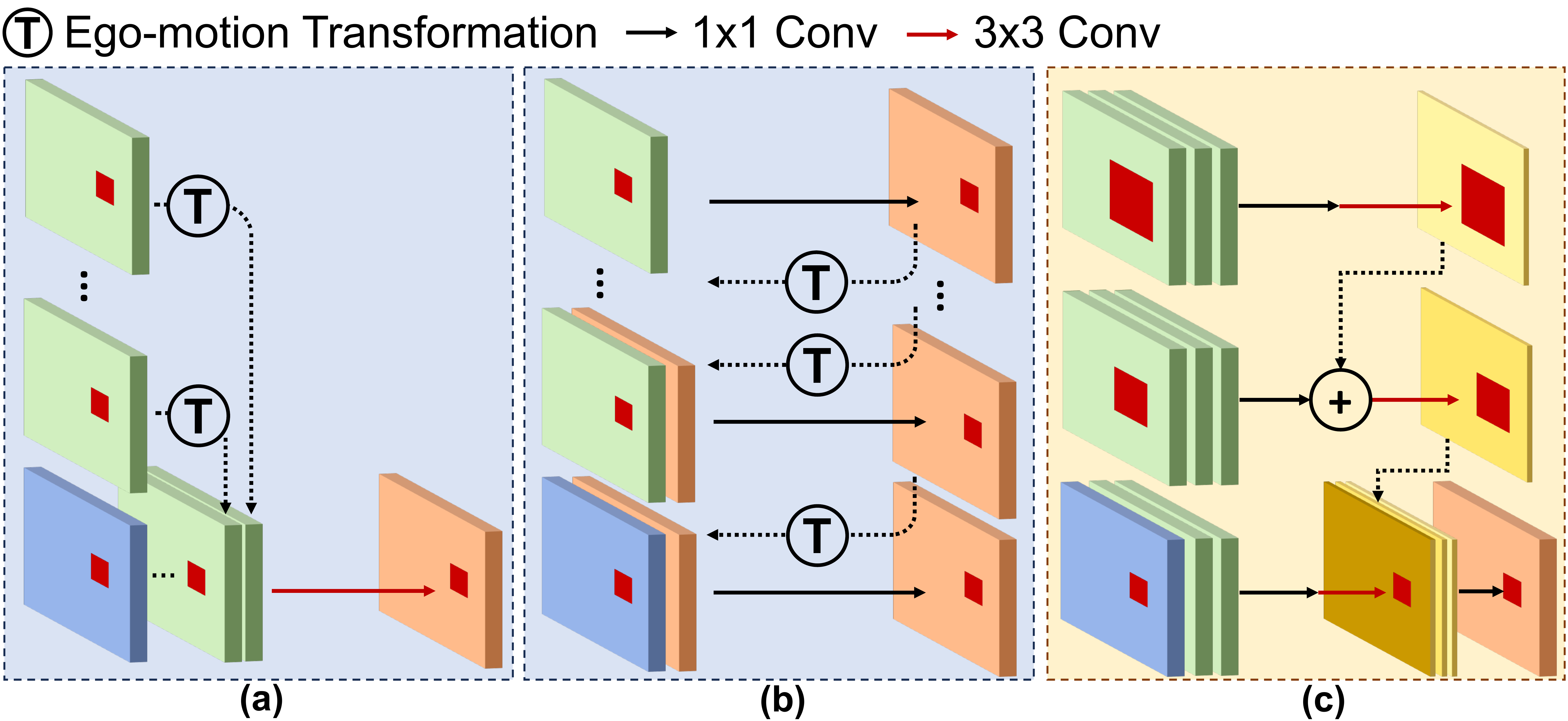}
    \caption{\textbf{Overview of Res2Fusion.} We list three major types of BEV temporal fusion techniques: (a) parallel fusion, (b) recurrent fusion, and (c) Res2Fusion (used in BEVNeXt).}
    \label{fig:res2fusion}
    \vspace{-0.1in}
\end{figure}

The dense BEV-based approaches fuse the current BEV representation $B_t$ with past representations over an extended period, which is vital for perceiving dynamic 3D scenes, especially over long temporal windows where object locations significantly change. However, expanding the receptive field in BEV space is challenging; simply increasing the kernel size causes excessive computation and risks overfitting in uniform 3D environments~\cite{chen2023largekernel3d}.

To address these issues, we develop a temporal aggregation technique named Res2Fusion as shown in Fig.~\ref{fig:res2fusion}, which enlarges the receptive field by incorporating multi-scale convolution blocks from the Res2Net architecture~\cite{gao2019res2net}.
Given $k$ historical BEV features $\{B_{t-k+1}, ..., B_t\}$ in which $B_t$ represents the BEV feature at the current frame, we first partition adjacent BEV features into $g=\frac{k}{w}$ groups with a fixed window size $w$, in which zero padding is used on the least recent group if $k \bmod{w} \neq 0$. The window size $w$ determines how much short-term locality the aggregation enjoys. After window partitioning, $1\times1$ convolutions $\{K_i^{1\times1}\}_{i=1}^{g}$ are used on these groups individually to reduce the channel size, which can be expressed as:\begin{equation}
    B'_i = K_i^{1\times1}([B_{t-(i+1)\times w};...;B_{t-i\times w}]) (i=0,...,g),
\end{equation} where $[\cdot;\cdot]$ represents the concatenation operation. Next, the multi-scale convolutions proposed by~\cite{gao2019res2net} is used as:
\begin{equation}
    B''_i= \begin{cases}
    K_i^{3\times3}(B'_i) & \text{if $i=g$;}~\\
    K_i^{3\times3}(B'_i+B'_{i+1}) & \text{if $0<i<g$;} \\
    B'_i & \text{if $i=0$.}
\end{cases}
\end{equation}
The increased receptive field, in turn, allows us to skip ego-motion transformation across historical BEVs, which avoids motion misalignment issues~\cite{huang2023leveraging} in previous techniques~\cite{park2022time, han2023exploring}. Finally, the output of the Res2Fusion module $\tilde{B}$ can be written as:\begin{equation}
    \tilde{B} = K_{final}^{1\times1}([B''_g;...;B''_0]),
\end{equation} which is further processed with strided layers with a similar structure and an FPN~\cite{lin2017feature} for multi-scale information following BEVDet~\cite{huang2021bevdet}.

\subsection{Object Decoder with Perspective Refinement}
With the unified BEV representation $\tilde{B}$ available, we apply a LiDAR-based 3D object detection head (\eg CenterPoint Head~\cite{yin2021center}) to $\tilde{B}$ for final detection. Nevertheless, forward projection distorts 2D features, leading to a discrete and sparse BEV representation, as observed in FB-BEV~\cite{li2023fbbev}. Thus, we aim to compensate for the distortion using perspective refinement before regressing BEV features of ROI (Regions of Interest) to object attributes.

In the object decoder, we follow CenterPoint~\cite{yin2021center} to calculate the object heatmap $H$ by applying a $3\times3$ convolution and a sigmoid function to the output of the BEV Encoder $\tilde{B}$. The heatmap $H$ contains $K$ channels, corresponding to the $K$ object classes. For attribute regression, we first sample the features $B^{center}=\{\tilde{B}_{x,y}|H_{x,y}>\tau\}$ from object centers in $\tilde{B}$ by employing a heatmap threshold $\tau=0.1$. As a regression head in CenterPoint typically consists of three convolution layers (one is shared among all heads), we expand $B^{center}$ to $B^{roi}$ by taking into account a $7\times7$ neighboring region of each element in $B^{center}$. Along with a set of learnable queries $\{Q_{x,y}\}$, the feature set $B^{roi}$ then goes through the perspective refinement process through a spatial cross-attention layer following~\cite{li2022bevformer}: 
\begin{flalign}
    &SCA(B^{roi}_{x,y}, F_{1/n})=\nonumber&&\\ 
    &\sum_{i=1}^{N}\sum_{j=1}^{N_{ref}}\mathcal{F}_d(B^{roi}_{x,y}+Q_{x,y}, \mathcal{P}_i(x,y,z_j), F^i_{1/n}),&&
\end{flalign}
where $\mathcal{F}_d$ is the deformable attention function~\cite{zhu2020deformable} and $\mathcal{P}_i(x,y,z_j)$ is a reference point lifted to height $z_j$. To introduce depth guidance, we embed the 2D features with CRF-modulated depth probabilities $\boldsymbol{\tilde{d^i}}$, which are object-consistent after exploiting color information: 
\begin{flalign}
    &SCA(B^{roi}_{x,y}, F_{1/n})=&&\\
    &\sum_{i=1}^{N}\sum_{j=1}^{N_{ref}}\mathcal{F}_d(B^{roi}_{x,y}+Q_{x,y}, \mathcal{P}_i(x,y,z_j), F^i_{1/n}+Mlp(\boldsymbol{\tilde{d^i}})).\nonumber&&
\end{flalign} 
Finally, regression heads from CenterPoint regress the refined feature set $B^{roi}$ to the final object attributes.

\section{Experiments}
\label{sec:experiments}

\subsection{Implementation Details}
\begin{table*}[t]

\centering
\small

\begin{tabular}{l|cc|cc|c|cccc}
\toprule
\textbf{Method} & \textbf{Backbone} & \textbf{Input Size} & \textbf{NDS$\uparrow$}   & \textbf{mAP$\uparrow$}   & \textbf{mATE$\downarrow$}  & \textbf{mASE$\downarrow$}  & \textbf{mAOE$\downarrow$}  & \textbf{mAVE$\downarrow$}  & \textbf{mAAE$\downarrow$} \\
\midrule
BEVDepth~\cite{li2023bevdepth} & ResNet50 & $256 \times 704$ & 0.475 & 0.351 & 0.639 & 0.267 & 0.479 & 0.428 & 0.198 \\
BEVPoolv2~\cite{huang2022bevpoolv2} & ResNet50 & $256 \times 704$ & 0.526 & 0.406 & 0.572 & 0.275 & 0.463 & 0.275 & 0.188 \\
SOLOFusion~\cite{park2022time} & ResNet50 & $256 \times 704$ & 0.534 & 0.427 & 0.567 & 0.274 & 0.511 & 0.252 & 0.181 \\
VideoBEV~\cite{han2023exploring} & ResNet50 & $256 \times 704$ & 0.535 & 0.422 & 0.564 & 0.276 & 0.440 & 0.286 & 0.198 \\
Sparse4Dv2~\cite{lin2023sparse4d} & ResNet50 & $256 \times 704$ & 0.539 & \textbf{0.439} & 0.598 & 0.270 & 0.475 & 0.282 & \textbf{0.179} \\
StreamPETR~\cite{wang2023streampetr} & ResNet50 & $256 \times 704$ & 0.540 & 0.432 & 0.581 & 0.272 & 0.413 & 0.295 & 0.195 \\
SparseBEV~\cite{liu2023sparsebev} & ResNet50 & $256 \times 704$ & 0.545 & 0.432 & 0.606 & 0.274 & \textbf{0.387} & \textbf{0.251} & 0.186 \\
\rowcolor{gray!20}
\textbf{BEVNeXt} & ResNet50 & $256 \times 704$ & \textbf{0.548} & 0.437 & \textbf{0.550} & \textbf{0.265} & 0.427 & 0.260 & 0.208 \\
\midrule
StreamPETR*~\cite{wang2023streampetr} & ResNet50 & $256 \times 704$ & 0.550 & 0.450 & 0.613 & 0.267 & 0.413 & 0.265 & 0.196 \\
SparseBEV*~\cite{liu2023sparsebev} & ResNet50 & $256 \times 704$ & 0.558 & 0.448 & 0.581 & 0.271 & \textbf{0.373} & \textbf{0.247} & \textbf{0.190} \\
\rowcolor{gray!20}
\textbf{BEVNeXt*} & ResNet50 & $256 \times 704$ & \textbf{0.560} & \textbf{0.456} & \textbf{0.530} & \textbf{0.264} & 0.424 & 0.252 & 0.206 \\
\midrule
BEVDepth~\cite{li2023bevdepth} & ResNet101 & $512 \times 1408$ & 0.535 & 0.412 & 0.565 & 0.266 & 0.358 & 0.331 & 0.190 \\
SOLOFusion~\cite{park2022time} & ResNet101 & $512 \times 1408$ & 0.582 & 0.483 & 0.503 & 0.264 & 0.381 & 0.246 & 0.207 \\
SparseBEV*~\cite{liu2023sparsebev} & ResNet101 & $512 \times 1408$ & 0.592 & 0.501 & 0.562 & 0.265 & 0.321 & 0.243 & 0.195 \\
StreamPETR*~\cite{wang2023streampetr} & ResNet101 & $512 \times 1408$ & 0.592 & 0.504 & 0.569 & 0.262 & \textbf{0.315} & 0.257 & 0.199 \\
Sparse4Dv2*~\cite{lin2023sparse4d} & ResNet101 & $512 \times 1408$ & 0.594 & 0.505 & 0.548 & 0.268 & 0.348 & 0.239 & \textbf{0.184} \\
Far3D*~\cite{jiang2023far3d} & ResNet101 & $512 \times 1408$ &  0.594 & \textbf{0.510} & 0.551 & \textbf{0.258} & 0.372 & \textbf{0.238} & 0.195 \\
\rowcolor{gray!20}
\textbf{BEVNeXt*} & ResNet101 & $512 \times 1408$ & \textbf{0.597} & 0.500 & \textbf{0.487} & 0.260 & 0.343 & 0.245 & 0.197 \\
\midrule
StreamPETR~\cite{wang2023streampetr} & ViT-L & $320 \times 800$ & 0.609 & 0.530 & 0.564 & \textbf{0.255} & \textbf{0.302} & 0.240 & 0.207 \\
\rowcolor{gray!20}
\textbf{BEVNeXt} & ViT-Adapter-L~\cite{chen2022vision} & $320 \times 800$ & \textbf{0.622} & \textbf{0.535} & \textbf{0.467} & 0.260 & 0.309 & \textbf{0.227} & \textbf{0.195} \\
\bottomrule
\end{tabular}

\caption{
\textbf{Comparison on the nuScenes val set.} ViT-L~\cite{dosovitskiy2020image} is pretrained on COCO~\cite{lin2014microsoft} and Objects365~\cite{shao2019objects365}, while ViT-Adapter-L~\cite{chen2022vision} is pretrained on DINOv2~\cite{oquab2023dinov2}. * The backbone benefits from perspective pretraining~\cite{wang2021fcos3d}.
}
\vspace{-0.05in}
\label{tab:val}
\end{table*}

Our BEVNeXt builds upon BEVPoolv2~\cite{huang2022bevpoolv2}, a dense BEV-based framework featuring an efficient forward projection technique and the camera-aware depth network from BEVDepth~\cite{li2023bevdepth}. 
BEVPoolv2 is also the baseline model in the following experiments.
As the default configuration, we employ ResNet50~\cite{he2016deep} as the image backbone, an input resolution of $256 \times 704$ for multi-view images, and a grid size of $128 \times 128$ for the BEV space. Only when we use larger backbones (\ie ResNet101~\cite{he2016deep}, Vit-Adapter-L~\cite{chen2022vision}, V2-99~\cite{lee2019energy}), the BEV resolution is increased to $256\times256$. To maximize CRF-modulation's effect, as mentioned in Sec.~\ref{subsec:crf}, the depth network operates on $F_{1/8}$ to produce more fine-grained depth probabilities given an input resolution of $256\times704$, while on $F_{1/16}$ under other circumstances. When considering the incorporation of historical temporal information, we calculate BEV features from the past 8 frames in addition to the current frame, organized into 3 BEV groups with a window size of 3 for Res2Fusion.

For the training setup, our models are trained with data augmentations for both BEV and images~\cite{huang2021bevdet}. The CBGS strategy~\cite{zhu2019class} is adopted for a duration of 12 epochs, with the first 2 epochs without temporal information~\cite{park2022time}. When employing ViT-Adapter-L and V2-99 as the image backbone, a shorter duration of 6 epochs is adopted to prevent over-fitting. We use the AdamW optimizer~\cite{loshchilov2017decoupled} and a total batch size of 64. The training process is consistent with BEVDepth~\cite{li2023bevdepth}, where the depth network and the detection head are optimized under supervision simultaneously.

\subsection{Datasets and Metrics}
\begin{table*}[t]

\centering
\small

\begin{tabular}{l|cc|cc|c|cccc}
\toprule
\textbf{Method} & \textbf{Backbone} & \textbf{Input Size} & \textbf{NDS$\uparrow$}   & \textbf{mAP$\uparrow$}   & \textbf{mATE$\downarrow$}  & \textbf{mASE$\downarrow$}  & \textbf{mAOE$\downarrow$}  & \textbf{mAVE$\downarrow$}  & \textbf{mAAE$\downarrow$}  \\
\midrule
BEVDepth~\cite{li2023bevdepth} & V2-99 & $640\times1600$ & 0.600 & 0.503 & 0.445 & 0.245 & 0.378 & 0.320 & 0.126 \\
BEVStereo~\cite{li2022bevstereo} & V2-99 & $640\times1600$ & 0.610 & 0.525 & 0.431 & 0.246 & 0.358 & 0.357 & 0.138 \\
SOLOFusion~\cite{park2022time} & ConvNeXt-B & $640\times1600$ & 0.619 & 0.540 & 0.453 & 0.257 & 0.376 & 0.276 & 0.148 \\
FB-BEV~\cite{liu2023sparsebev} & V2-99 & $640\times1600$ & 0.624 & 0.537 & 0.439 & 0.250 & 0.358 & 0.270 & 0.128 \\
VideoBEV~\cite{han2023exploring} & ConvNeXt-B & $640\times1600$ & 0.629 & 0.554 & 0.457 & 0.249 & 0.381 & 0.266 & 0.132 \\
BEVFormerv2~\cite{yang2023bevformer} & InternImage-XL~\cite{wang2023internimage} & $640\times1600$ & 0.634 & 0.556 & 0.456 & 0.248 & \textbf{0.317} & 0.293 & 0.123 \\
StreamPETR~\cite{wang2023streampetr} & V2-99 & $640\times1600$  & 0.636 & 0.550 & 0.479 & 0.239 & \textbf{0.317} & 0.241 & 0.119 \\
SparseBEV~\cite{liu2023sparsebev} & V2-99 & $640\times1600$  & 0.636 & 0.556 & 0.485 & 0.244 & 0.332 & 0.246 & 0.117 \\
Sparse4Dv2~\cite{lin2023sparse4d} & V2-99 & $640\times1600$ & 0.638 & 0.556 & 0.462 & \textbf{0.238} & 0.328 & 0.264 & \textbf{0.115} \\
\rowcolor{gray!20}
\textbf{BEVNeXt} & V2-99 & $640\times1600$ & \textbf{0.642} & \textbf{0.557} & \textbf{0.409} & 0.241 & 0.352 & \textbf{0.233} & 0.129 \\
\bottomrule
\end{tabular}

\caption{
\textbf{Comparison on the nuScenes test set.} ConvNeXt-B~\cite{liu2022convnet} is pretrained on ImageNet-22K~\cite{deng2009imagenet}, while V2-99 is initialized from a DD3D~\cite{park2021pseudo} backbone. The listed methods do not use future frames during training or testing.
}
\vspace{-0.05in}
\label{tab:test}
\end{table*}

\begin{table}[h]
  \centering
  \vspace{-1em}
  \small
  \setlength{\tabcolsep}{0pt} 
  \resizebox{\linewidth}{!}{
  \begin{tabular*}{\linewidth}{@{\extracolsep{\fill}}lcccccccc@{}}
    \toprule
\textbf{Method} & \textbf{AMOTA$\uparrow$}   & \textbf{AMOTP$\downarrow$}  & \textbf{RECALL$\uparrow$}  & \textbf{IDS$\downarrow$} \\
\midrule
SPTR-QTrack~\cite{wang2023streampetr, yang2022quality} & 0.566 & 0.975 & 0.650 & 784  \\
DORT~\cite{qing2023dort} & 0.576 & 0.951 & 0.634 & 774  \\
\textbf{BEVNeXt-PolyMot~\cite{li2023poly}} & \textbf{0.578} & \textbf{0.917} & \textbf{0.720} & \textbf{519} \\
\bottomrule
  \end{tabular*}
  }
  \caption{
\textbf{3D multi-object tracking on the nuScenes test set.} Ours uses V2-99 as the backbone while the others use ConvNeXt-B.
}
\vspace{-2em}
\label{tab:track}
\end{table}
We conduct extensive evaluations of our method on nuScenes~\cite{caesar2020nuscenes}, a multi-modal dataset encompassing 1000 distinct scenarios, which are recorded with a 32-beam LiDAR, six surround-view cameras, and five radars, annotated at a 2Hz frequency. Our assessment is based on the metrics of nuScenes, including mean Average Precision (mAP), mean Average Translation Error (mATE), mean Average Scale Error (mASE), mean Average Orientation Error (mAOE), mean Average Velocity Error (mAVE), mean Average Attribute Error (mAAE), and the final nuScenes Detection Score (NDS). 3D Multi-Object Tracking is evaluated on Average Multi-Object Tracking Accuracy (AMOTA), Average Multi-Object Tracking Precision (AMOTP), Recall, and ID Switch (IDS).

\subsection{Main Results}

\begin{table*}[t]

\centering
\small

\begin{tabular}{l|cc|c|cc|ccc}
\toprule
\textbf{Backbone} & \textbf{Input Size} & \textbf{LiDAR Coverage} & \textbf{CRF} & \textbf{NDS$\uparrow$} & \textbf{mAP$\uparrow$} & \textbf{mATE$\downarrow$} & \textbf{mAOE$\downarrow$} & \textbf{mAVE$\downarrow$}\\
\midrule
\multirow{4}{*}{ResNet50} & \multirow{2}{*}{$256 \times 704$} & \multirow{2}{*}{$\approx 85\%$} &  & 0.490 & 0.368 & 0.628 & \textbf{0.485} & \textbf{0.355}\\
 &  &  & \checkmark & \textbf{0.492} & \textbf{0.372} & \textbf{0.614} & 0.490 & 0.369\\
\cmidrule{2-9}
 & \multirow{2}{*}{$384 \times 1056$} & \multirow{2}{*}{$\approx 65\%$} &  & 0.502 & 0.390 & 0.599 & 0.456 & 0.395\\
 &  &  & \checkmark & \textbf{0.521} & \textbf{0.406} & \textbf{0.581} & \textbf{0.419} & \textbf{0.343}\\
\cmidrule{1-9}
\multirow{2}{*}{ResNet101} & \multirow{2}{*}{$512 \times 1408$} & \multirow{2}{*}{$\approx 50\%$} &  & 0.535 & 0.412 & 0.565 & 0.358 & 0.331\\
 & & & \checkmark & \textbf{0.553} & \textbf{0.433} & \textbf{0.544} & \textbf{0.332} & \textbf{0.308}\\
\bottomrule
\end{tabular}

\caption{
\textbf{Ablation of CRF modulation with different backbones and input resolutions.} All depth networks operate on $F_{1/16}$. Only 1 history frame is used. The effect of CRF modulation is minor given dense point clouds supervision.
}
\vspace{-0.1in}
\label{tab:abl_crf}
\end{table*}

Tab.~\ref{tab:val} shows detailed detection performance on the validation split of nuScenes. Specifically, using a ResNet50~\cite{he2016deep} backbone and an input resolution of $256\times704$, BEVNeXt outperforms BEV-based SOLOFusion~\cite{park2022time} by a clear margin of 2.6\% NDS and 2.9\% mAP, exceeding recently proposed query-based SparseBEV~\cite{liu2023sparsebev} with and without perspective pretraining~\cite{wang2021fcos3d}.
This demonstrates the detection abilities of BEVNeXt in a lightweight setting.
Furthermore, equipped with ResNet101, BEVNeXt outperforms the strong query-based framework Far3D~\cite{jiang2023far3d} by 0.3\% NDS, while using  ViT-Adapter-L~\cite{chen2022vision}, BEVNeXt yields a new state-of-the-art result of 62.2\% NDS on the val split.
These experiments verify the scalability of our proposed modules, which allows BEVNeXt to thrive with a larger and more modern backbone.
On the test split of nuScenes, Tab.~\ref{tab:test} shows that BEVNeXt outperforms all prior methods with a moderate-sized V2-99~\cite{lee2019energy} backbone under the same input resolution of $640\times1600$. 
Specifically, BEVNeXt surpasses the BEV-based SOLOFusion~\cite{park2022time} by a 2.3\% NDS and the query-based Sparse4Dv2~\cite{lin2023sparse4d} by a 0.4\% NDS. 
Meanwhile, BEVNeXt produces fewer translation errors (\ie mATE) than prior detectors consistently, demonstrating its superiority in object localization.
In particular, BEVNeXt achieves 5.3\% less mATE compared with Sparse4Dv2~\cite{lin2023sparse4d}.
Integrating the PolyMot~\cite{li2023poly} tracker with the detections of BEVNeXt-V2-99, BEVNeXt surpasses the BEV-based specialized tracker DORT by 0.2\% AMOTA on the nuScenes test set (Tab.~\ref{tab:track}).

\subsection{Ablation Studies}
\begin{table}[h]
\vspace{-0.1in}

\centering
\resizebox{\columnwidth}{!}{%

\begin{tabular}{cccc|cc}
\toprule
\textbf{Res2Fusion} & \textbf{$F_{1/n}$} & \textbf{CRF} & \textbf{Refinement} & \textbf{NDS$\uparrow$} & \textbf{mAP$\uparrow$}\\
\midrule
& $16$ & & & 0.526 & 0.406\\
\checkmark & $16$ & & & 0.537 & 0.420\\
\checkmark & $8$ & & & 0.540 & 0.430\\
\checkmark & $8$ & \checkmark & & 0.542 & 0.434\\
\checkmark & $8$ & \checkmark & \checkmark & \textbf{0.548} & \textbf{0.437}\\

\bottomrule
\end{tabular}

}
\caption{
\textbf{Ablation of BEVNeXt Components}. The baseline is BEVPoolv2 with an input resolution of $256\times704$ , ResNet50 as the backbone, and a long-term history of 8 frames.
}
\vspace{-0.1in}
\label{tab:abl_component}
\end{table}

\vspace{0.05in}\noindent\textbf{Ablation on Different Components.} To verify the effectiveness of our proposed components, we gradually remove the components of BEVNeXt. 
As depicted in Tab.~\ref{tab:abl_component}, the absence of each component causes a distinct degradation of NDS and mAP. 
Though the increase of the feature map scale aims to accommodate CRF modulation, we observe this modification brings performance improvements (+0.3\% NDS) itself, which is also confirmed in the localization potential analysis made by SOLOFusion~\cite{park2022time}.
The integrated version of BEVNeXt brings an overall performance gain of 2.2\% NDS and 3.1\% mAP compared with the baseline.

\vspace{0.05in}\noindent\textbf{CRF Modulation.} In Tab.~\ref{tab:abl_crf}, the role of CRF modulation under sparse supervision is studied. The impact of CRF modulation is insignificant (+0.2\% NDS) when supervision is dense already, but becomes much more notable (+1.9\% NDS) under sparse supervision.
It can be shown that the object-level consistency provided by CRF modulation comprehensively enhances the detection performance of BEV frameworks, which also demonstrates scalability (+1.8\% NDS) using a larger backbone (\ie ResNet101).
Yet, in long-term settings, the improvements become moderate due to saturating localization potential~\cite{park2022time}, as displayed in Tab.~\ref{tab:abl_component}.

\begin{table}[t]

\centering
\small
\resizebox{\columnwidth}{!}{

\begin{tabular}{cc|cc}
\toprule
\textbf{Ego-motion Trans.} & \textbf{Window Size $w$} & \textbf{NDS$\uparrow$} & \textbf{mAP$\uparrow$} \\
\midrule
- & 2 & 0.531 & 0.417\\
\checkmark & 3 & 0.535 & 0.416\\
- & 3 & \textbf{0.537} & \textbf{0.420}\\
- & 4 & 0.529 & 0.418\\
\bottomrule
\end{tabular}

}
\caption{
\textbf{Ablation of Res2Fusion.} We compare different window sizes $w$ and the effect of ego-motion transformation over 8 historical frames (9 frames in total). Zero padding is used if the number of frames cannot be divided evenly by $w$.
}
\vspace{-0.1in}
\label{tab:abl_res2fusion}
\end{table}
\begin{table}[h]

\centering
\resizebox{\columnwidth}{!}{%

\begin{tabular}{cc|cc}
\toprule
\textbf{Depth Embedding} & \textbf{Depth Source} & \textbf{NDS$\uparrow$} & \textbf{mAP$\uparrow$}\\
\midrule
- & - & 0.497 & 0.379\\
\checkmark & Depth Network & 0.501 & 0.379\\
\checkmark & CRF & \textbf{0.505} & \textbf{0.382}\\

\bottomrule
\end{tabular}

}
\caption{
\textbf{Ablation of Depth Embedding in Perspective Refinement.} All depth networks operate on $F_{1/8}$, as the input resolution is $256\times704$. Only 1 history frame is used.
}
\vspace{-0.1in}
\label{tab:abl_depthemb}
\end{table}

\vspace{0.05in}\noindent\textbf{Design of Res2Fusion.} Tab.~\ref{tab:abl_res2fusion} shows how Res2Fusion is affected by the window size and ego-motion transformation. The window size determines the number of adjacent BEVs to process in parallel, which only demands smaller receptive fields than the long-term scenario.
Our experiment shows that a window size of 3 maximizes the effect of Res2Fusion by reaching a balance between the short-term locality and the long-term receptive field.
In addition, forcibly warping previous BEV features to the current timestamp causes a performance degradation, which is caused by the misalignment of dynamic objects~\cite{huang2023leveraging}.

\vspace{0.05in}\noindent\textbf{Effect of Depth Embedding.} In Tab.~\ref{tab:abl_depthemb}, we ablate the effect of depth embedding in the perspective refinement module. 
The purpose of depth embedding is to help the 3D object decoder attend to discriminative features by forming object-level consistencies in the 2D space.
When the depth is produced by the depth network, there exists positive but minor influences. However, when the CRF-modulated depth information is adopted, we witness an 0.8\% increase in NDS, which is a boost in the prediction of object attributes.
This phenomenon also verifies the effect of CRF modulation.

\subsection{Visualization and Efficiency Analysis}
\begin{table}[t]
\vspace{-0.1in}
\centering
\small
\resizebox{\columnwidth}{!}{

\begin{tabular}{lc|cc}
\toprule
\textbf{Component} & \textbf{Enhancements} & \textbf{Param. (M)} & \textbf{GFLOPs} \\
\midrule
\multirow{2}{*}{Depth Network} & w/o CRF & 27.3 & 3864.4 \\
& w/ CRF & 27.3 & 3873.9 \\
\midrule
\multirow{2}{*}{BEV Encoder} & Parallel Fusion & 54.6 & 172.8 \\
& Res2Fusion & 6.9 & 31.4 \\
\midrule
\multirow{2}{*}{Detection Head} & w/o Refinement & 1.6 & 99.7 \\
& w/ Refinement & 2.3 & 125.4 \\
\toprule

\multicolumn{2}{c|}{\textbf{Method}} & \multicolumn{2}{c}{\textbf{Speed (FPS)}}\\
\midrule
\multicolumn{2}{c|}{StreamPETR-R101~\cite{wang2023streampetr}} & \multicolumn{2}{c}{6.4} \\
\multicolumn{2}{c|}{SOLOFusion-R101~\cite{park2022time}} & \multicolumn{2}{c}{1.5} \\
\multicolumn{2}{c|}{BEVNeXt-R101} & \multicolumn{2}{c}{4.4} \\

\bottomrule
\end{tabular}

}
\vspace{-0.1in}
\caption{
\textbf{Analysis of Runtime Efficiency}. The listed methods use ResNet101 as the image backbone. Both SOLOFusion-R101 and BEVNeXt-R101 utilize a BEV resolution of $256\times256$.
}
\vspace{-0.2in}
\label{tab:performance}
\end{table}
\begin{figure}[h]
    \centering
    \vspace{-0.15in}
    \includegraphics[width=\linewidth]{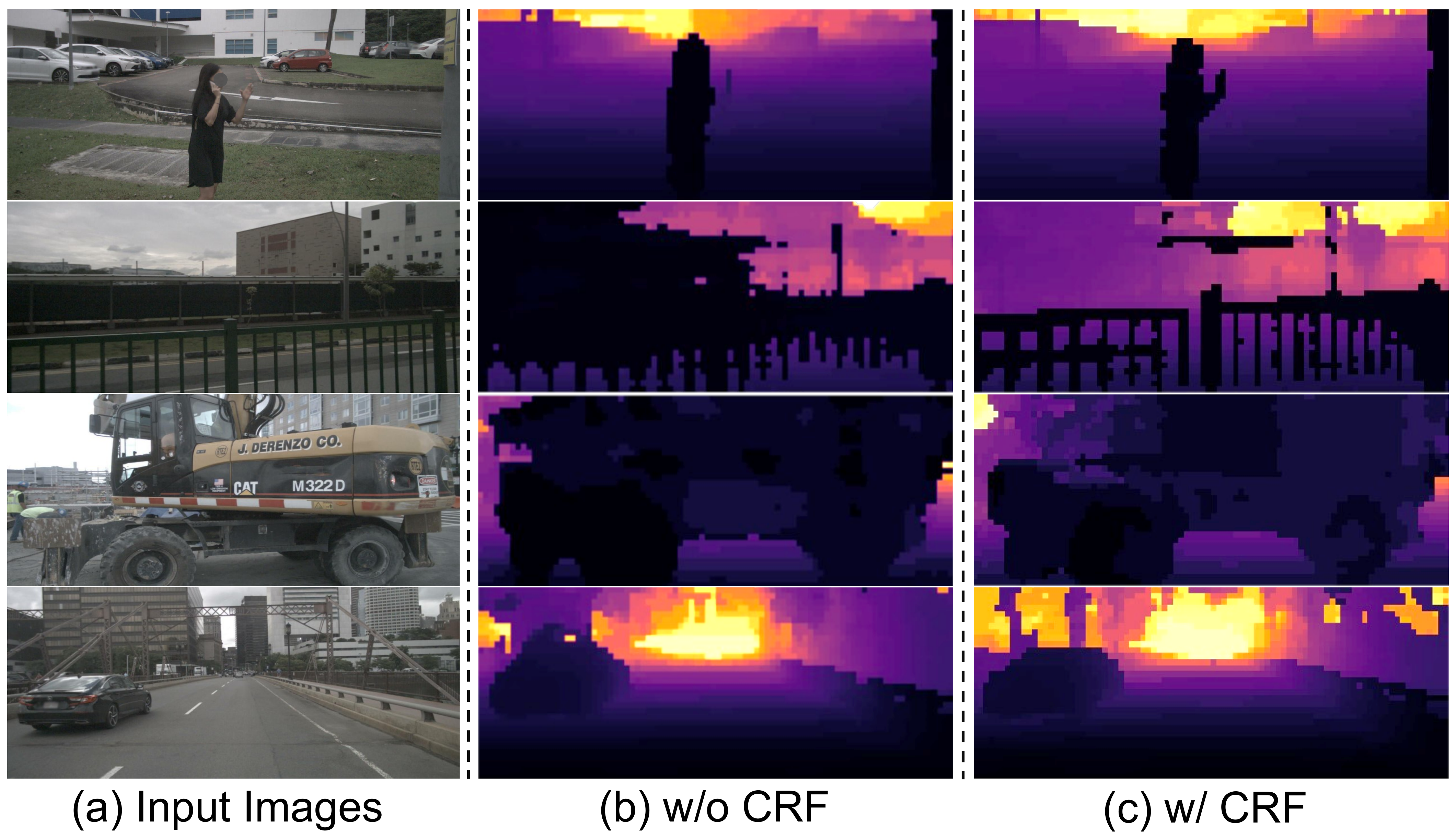}
    \vspace{-0.2in}
    \caption{\textbf{Comparison of Depth Estimation with and without CRF modulation on the nuScenes val split.} We visualize depth ranges using an argmax operation on various depth bins. The CRF-modulated depth probabilities can distinguish objects from the background better.}
    \label{fig:vis_crf}
    \vspace{-0.15in}
\end{figure}

\begin{figure}[h]
    \centering
    \vspace{-0.1in}
    \includegraphics[width=\linewidth]{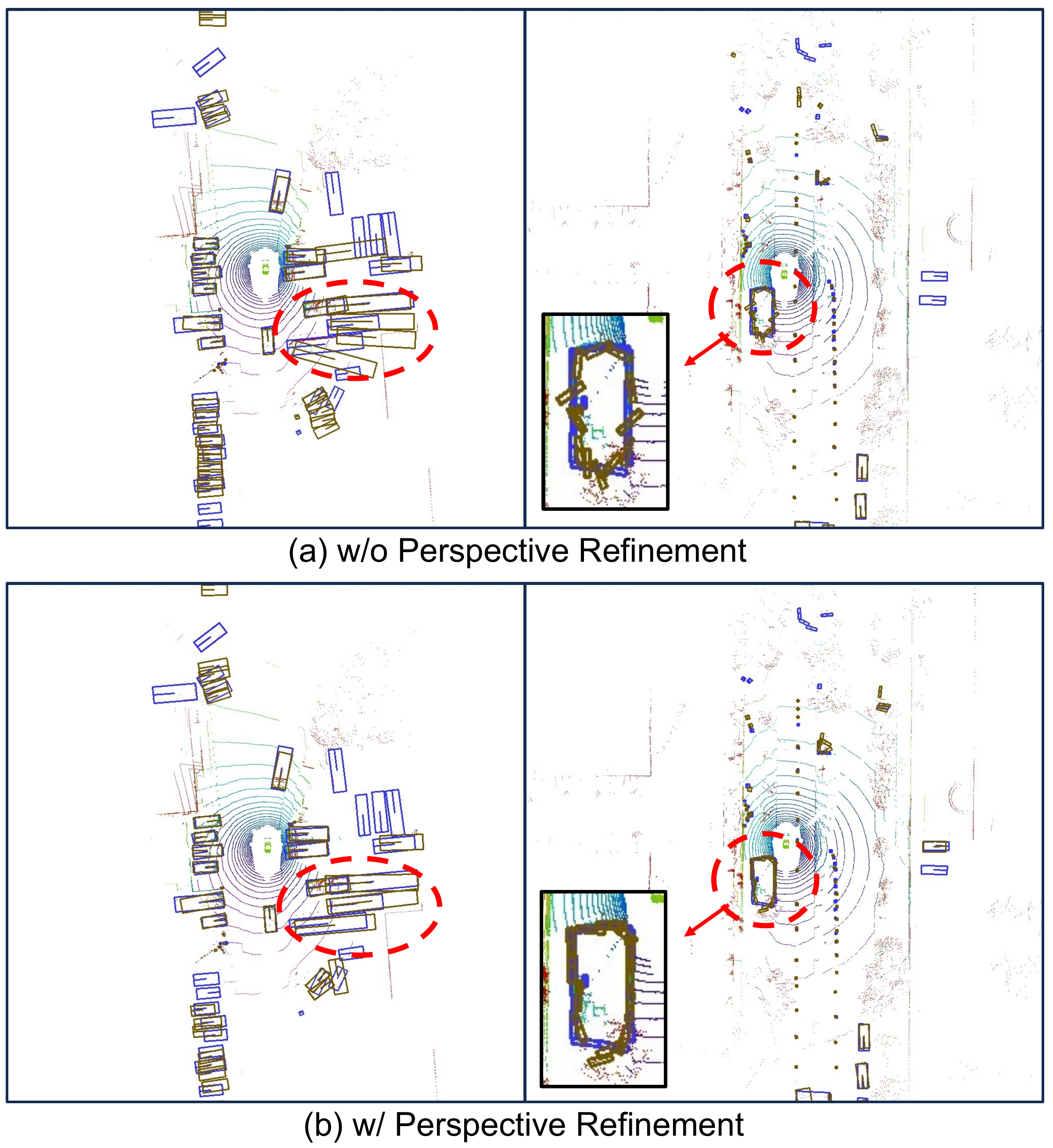}
    \vspace{-0.2in}

    \caption{\textbf{Comparison of Detection Results with and without Perspective Refinement on the nuScenes val split.} Compared with the coarse predictions of a CenterPoint head~\cite{yin2021center}, our refined objects are more aligned with the ground truths.}
    \label{fig:vis_pers}
    \vspace{-0.2in}

\end{figure}

\vspace{0.05in}\noindent\textbf{Visualization.} We first visualize CRF-modulated depth estimation in Fig.~\ref{fig:vis_crf}. 
The modulated depth probabilities are more boundary-sticky and achieve higher object-level consistencies. 
Besides, they contain fewer artifacts, which interfere with the spatial accuracy of BEV features. 
For perspective refinement, both large and small objects can benefit from this process as shown in Fig.~\ref{fig:vis_pers}. Compared with a coarse object decoder, refined objects are more accurate in orientation.

\vspace{0.05in}\noindent\textbf{Efficiency Analysis.} As shown in Tab.~\ref{tab:performance}, our proposed modules require negligible or even less computation, making them suitable for deployment. With a PyTorch fp32 backend and an RTX 3090, though falling behind query-based StreamPETR~\cite{wang2023streampetr}, BEVNeXt is faster than BEV-based SOLOFusion~\cite{park2022time}. This can be attributed to the absence of a temporal stereo, which is computationally expensive.

\section{Conclusion}
\label{sec:conclusion}
In this work, we proposed a fully enhanced dense BEV framework for multi-view 3D object detection dubbed BEVNeXt. 
We first identify three shortcomings of classic dense BEV-based frameworks: (1) insufficient 2D modeling, (2) inadequate temporal modeling, and (3) feature distortion in uplifting.
To address these inherent issues, we propose three corresponding components: (1) CRF-modulated depth estimation, (2) Res2Fusion for long-term temporal aggregation, and (3) an object decoder with perspective refinement.
Extensive experiments are carried out, showing that BEVNeXt excels in object localization and supersedes sparse query paradigms and dense BEV frameworks on the nuScenes benchmark. 
Specifically, BEVNeXt achieves a new state-of-the-art of 56.0\% NDS and 64.2\% NDS on the nuScenes val split and test split, respectively.

\vspace{0.05in}\noindent\textbf{Limitations.} Though BEVNeXt demonstrates stronger performance than existing sparse query paradigms, it still falls behind in terms of efficiency. Integrating BEV frameworks into long-range settings also present challenges. We expect these issues to be addressed in future research.

\noindent\textbf{Acknowledgement} This project was supported by NSFC under Grant No. 62102092.

{\small
\bibliographystyle{ieeenat_fullname}
\bibliography{11_references}
}


\end{document}


\title{\paperTitle}
\author{\authorBlock}
\maketitlesupplementary

\section{Appendix Section}
Supplementary material goes here.

{\small
\bibliographystyle{ieee_fullname}
\bibliography{11_references}
}